\documentclass[conference]{IEEEtran}
\usepackage{times}

\usepackage[numbers]{natbib}
\usepackage{multicol}
\usepackage[bookmarks=true]{hyperref}
\usepackage{graphicx}
\usepackage{subcaption}
\usepackage{amssymb}
\usepackage{amsmath}
\usepackage{hyperref}

\let\oldtwocolumn\twocolumn
\renewcommand\twocolumn[1][]{%
    \oldtwocolumn[{#1}{
    \begin{center}
       \vspace{-1.cm}
       \includegraphics[width=\textwidth]{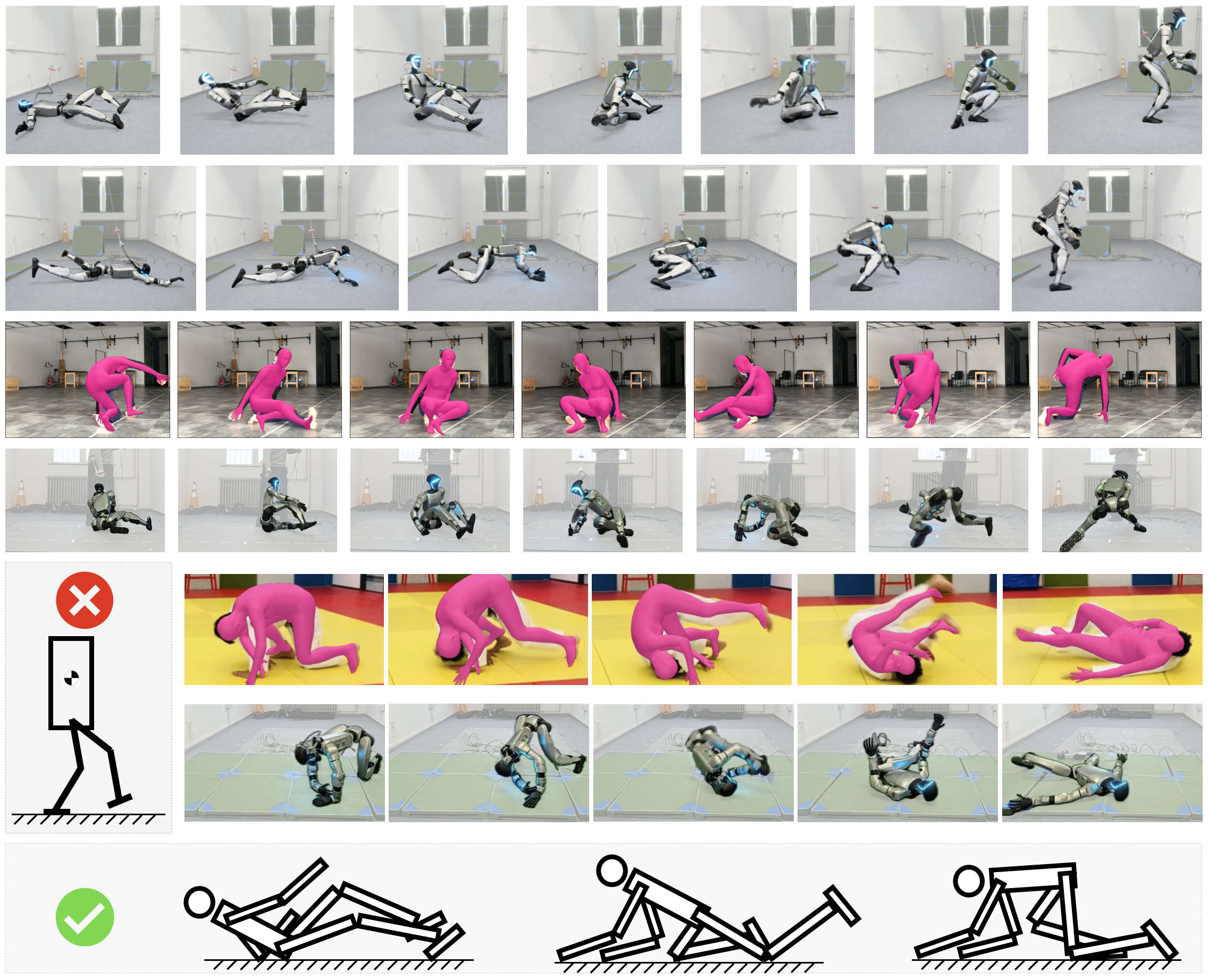}
       \captionof{figure}{We present a unified humanoid motion interface and a zero-shot sim-to-real reinforcement learning framework, so that humanoid robots can successfully perform extreme contact-agnostic motion in the real world. More detail at \href{https://project-instinct.github.io}{https://project-instinct.github.io}}
       \label{fig:teaser}
       \vspace{-0.1cm}
    \end{center}
    }]
}

\begin{document}

\title{Embrace Collisions: Humanoid Shadowing for Deployable Contact-Agnostics Motions}


\author{\authorblockN{Ziwen Zhuang}
\authorblockA{IIIS, Tsinghua University\\
Shanghai Qi Zhi Institute}
\and
\authorblockN{Hang Zhao}
\authorblockA{IIIS, Tsinghua University\\
Shanghai Qi Zhi Institute}
}


%

\maketitle
\begin{abstract}
Previous humanoid robot research works treat the robot as a bipedal mobile manipulation platform, where only the feet and hands contact the environment. However, we humans use all body parts to interact with the world, \textit{e.g.}, we sit in chairs, get up from the ground, or roll on the floor.
Contacting the environment using body parts other than feet and hands brings significant challenges in both model-predictive control and reinforcement learning-based methods. An unpredictable contact sequence makes it almost impossible for model-predictive control to plan ahead in real time. The success of the zero-shot sim-to-real reinforcement learning method for humanoids heavily depends on the acceleration of GPU-based rigid-body physical simulator and simplification of the collision detection.
Lacking extreme torso movement of the humanoid research makes all other components non-trivial to design, such as termination conditions, motion commands and reward designs. To address these potential challenges, we propose a general humanoid motion framework that takes discrete motion commands and controls the robot's motor action in real time. Using a GPU-accelerated rigid-body simulator, we train a humanoid whole-body control policy that follows the high-level motion command in the real world in real time, even with stochastic contacts and extremely large robot base rotation and not-so-feasible motion command.
\end{abstract}

\IEEEpeerreviewmaketitle

\section{Introduction}
Humans do not only walk and manipulate objects—they sit, lie down, get up from the ground, and transition between various postures. However, contemporary humanoid research largely defines humanoid robots as bipedal mobile manipulation platforms, focusing on walking and dexterous hand use while neglecting the full range of human motion. This limited perspective overlooks the advantages of humanoid morphology in whole-body locomotion and control.
This work enables humanoid robots to execute motions that are difficult or impossible for other morphologies, such as getting up from lying down, recovering from a prone position, or transitioning through kneeling postures.

Achieving such a set of general humanoid motions presents significant challenges: (1) designing an effective motion command interface across a wide range of postures, and (2) developing a deployable humanoid control policy that can be trained in simulation.

\subsection{Challenges in Motion Command Interface}
Previous locomotion controllers define motion commands in terms of horizontal movement, typically using velocity commands (e.g., forward-lateral speed, heading angular velocity) or waypoint sequences that guide the robot over short time horizons.
For example, when a quadruped robot stands from a crawling pose, the linear velocity command switches from the positive x-axis of the base frame to the negative z-axis of the base frame \cite{lib2024learning}.
However, such representations become ambiguous when the robot undergoes significant roll or pitch rotations. When commanding a humanoid to switch between standing and crawling, defining movement as linear velocity in the base frame becomes impractical, requiring extensive manual coding.


Furthermore, interfacing between high-level plans and real-time control presents additional challenges. One option is to use motion-tracking commands that directly follow joint positions \cite{he2024learning}, but this approach forces the high-level model to operate in real-time, which is computationally demanding. Alternatively, waypoint-based locomotion planning introduces potential odometry drift over long horizons.


Unlike locomotion commands, general motion commands lack a straightforward parameterization~\cite{zhuang2024humanoid, ilija2024humanoid, cheng2024express, bart2024robustSaW, gu2024humanoid, long2024learninghumanoidlocomotionperceptive, margolis2022walktheseways}. 
The high-dimensional sampling space makes it difficult to generate feasible motions without extensive filtering. A more reasonable option is to use human motion datasets as references for training humanoid policies. However, commonly used datasets such as AMASS~\cite{mahmood2019amass} predominantly contain standing postures, limiting their applicability to extreme motion scenarios. Additionally, kinematic differences between humans and humanoid robots cause many recorded motions to be infeasible. To address this, we construct an extreme-action dataset that includes motions with rich contact interactions.

\subsection{Challenges in Producing a Deployable Humanoid Control Policy}

Deploying complex humanoid motions on real hardware faces multiple constraints. Most recent works on reinforcement learning (RL)-based locomotion~\cite{zhuang2024humanoid, ilija2024humanoid, cheng2024express, bart2024robustSaW, gu2024humanoid, long2024learninghumanoidlocomotionperceptive, zhang2024wholebody, serifi2024vmp} rely on GPU-accelerated rigid-body simulation to train robust policies that can be transferred to real world. However, they only focus on bipedal locomotion, assuming foot-only contact with the ground, upper-body interactions are typically ignored. 
To enable general contact interactions, determining the optimal level of collision simplification in simulation remains a non-trivial question.

Apart from reinforcement learning, model-based control methods also simplify the robot model when planning the contact sequences. \citet{Khatib2008AUF, matthew2021MIThumanoids, nelson2019Atlas} assume only feet are contacting the ground and optimize the control function with a manually built kinematic model. Even in quadruped locomotion, motion planners divide movement into swing and contact phases \cite{gerardo2018MITcheetah}, significantly constraining possible motions. Introducing additional future contacts brings whole-body motion planning to the next level of difficulty, not to mention optimizing a collection of perception, state estimation, planning, and control systems that run accurately and in real time.


In this work, we propose a general framework that enables humanoid robots to execute a broad spectrum of extreme motions.
Specifically, we express all motion targets in the robot base frame, no matter whether the robot is standing or lying down. We use a key-frame-based method to express future motion targets, so that the robot receives information of future motion expectations. To investigate the possibility of training such an unusual humanoid motion on a real robot, we build an extreme-action dataset where the torsos are in extreme roll and pitch orientations. With this training pipeline, we show that using the simplified collision shape of the robot and domain randomization techniques, the control policy trained only in simulation can successfully follow the target motion trajectory in the real world.
We identify 3 technical designs to achieve such challenging motions:
\paragraph{Transformer-based motion command encoder with key-frame-based motion command}
To meet the need of varying the number of motion command inputs, we adopt a multi-headed self-attention network as our encoder for the motion command encoder. We use a transformer-based encoder block to encode the joint position command, link position command, and the base pose command in a sequence, as well as the error of the robot's current state with respect to the specific motion target.

\paragraph{Advantage mixing to bridge the gap of sparse motion reward and dense regularization reward}
To perform these extremely difficult contact-agnostic motions, the motion command sequences sampled from the existing dataset are not physically feasible for the given robot. The motion command has to be sparse and leave enough space for the policy to explore and generate solutions through reinforcement learning training. This task setup forces the task reward to be sparse throughout each episode. However, to protect the hardware when running, hardware-safe reward terms are typically dense, for example minimizing energy consumption, minimizing torque output, minimizing joint acceleration, etc. The task signal could be wrongly captured by the neural network. Thus we use a multi-critic technique called advantage mixing to bridge the gap between the sparse motion task reward and the dense regularization reward.

\paragraph{Newly designed termination condition to fit all base rotation conditions}
Previous works on humanoid locomotion and imitation considers only the standing condition. So, generally, when the robot's base orientation is greater than a certain threshold or the robot's base height is below a certain threshold, the training rollouts can be terminated. This is not the case when defining motion tasks, such as spinning on the ground or getting up from the ground. The designed working condition for the humanoid robot is already the ``failure condition'' in previous work. Thus, we redefined the termination conditions when the robot's current state deviates from the motion target too much when the motion target is expected to be reached.

\section{Related Works}

\paragraph{Humanoid Whole-Body Control}
Whole-body control for robots with multiple parts is a long-standing challenging problem. For humanoid robots, whole-body control with only feet contacting the ground is already a state-of-the-art research topic, due to the number of multiple rigid bodies attached on the system. Traditional methods depend on modeling the dynamics of the entire humanoid skeleton~\cite{matthew2021MIThumanoids, behzad2008whole, grizzle2009mabel, Moro2019whole, antonin2023synchronized, kajita2001the}. To simplify the computation and deploy the algorithm onboard in real-time, model-based methods use an inverted pendulum model to plan the footstep positions and then use whole-body control methods to acquire the targets for each motor, but without any additional contacts with the environment~\cite{Moro2019whole, kajita2001the}. However, these methods significantly limit the possibility of contacting the environments with components other than feet. For example, bending down and crawling on the floor introduces more contact such as knees and hands. These contacts can be unpredictable or cannot be predicted accurately as the model predictive control requires.

In learning-based methods, whole-body control is still a challenging topic. Most of them can be viewed as a combination of lower-body and upper-body~\cite{lu2024pmp, fu2024humanplus, cheng2024express, gu2024humanoid}. More integrated humanoid algorithms control all the joints on the humanoid robot but they only involve the contacts of feet~\cite{zhuang2024humanoid, ilija2024humanoid, serifi2024vmp, long2024learninghumanoidlocomotionperceptive, he2024learning, gu2024humanoid} or the contacts of feet and hands~\cite{zhang2024wococo, he2024omnih2o}. These workspaces significantly limit the potential of humanoid morphology. Unexpected contacts will make the robot significantly deviate from its original position and lead to further chaos. For example, a humanoid robot can sit on the driver's seat at emergency, but the contact of the humanoid's hip is not considered in this previous humanoid whole-body control research. Also, when dealing with skills on the floor, the current learning-based framework does not take knee, torso, or elbow collisions with the ground into consideration. This does not distinguish humanoid robots from quadruped robots if manipulators are attached~\cite{ma2022combining, zhang2024learning}.

\paragraph{General Motion Interface for Humanoid}
To design a unified interface for large models to control the humanoid robot without real-time requirements, whole-body control algorithms for humanoid robots design the interface depending on how the general motion is defined. For mobile manipulation tasks, the general interface can be described as a locomotion goal and upper-body joint position goal, such as~\cite{zhuang2023robot, zhuang2024humanoid, cheng2024express}. Specifically, locomotion can also be defined as a short-range navigation task~\cite{lee2020learning, yang2021real, miki2024learning}. However, all these interfaces intrinsically limit the potential of more complex behaviors, resulting in less flexibility to meet the fine-detailed motion targets.

In humanoid motion generation research, \citet{peng2018deepMimic, tessler2023calm, Luo2023PerpetualHC} defines all joint orientation and base position sequences as the interface. However, the imitation target is too strong. The robot has to follow the motion target at each timestep, leading to less tolerance when the motion target is not physically feasible for the current robot model. Although this interface can be applied to the real robot system~\cite{he2024learning, fu2024humanplus, he2024omnih2o}, this instant motion target still has to meet the real-time requirement. To deal with this issue, \citet{zargarbashi2024robotkeyframing, tessler2024maskedmimic, zhang2024wococo} propose using a sequence of future motion targets for the robot's control policy to follow. Before the motion sequence is exhausted, the system that generates motion targets can run in non-real-time mode. However, these motion frames introduce global odometry, which introduces accumulated error as the system runs between the refresh of high-level motion commands. These works all use a global odometry system to track the accurate motion of the robot, which is expensive to acquire and deploy in a user-oriented environment. To what extent the accumulated error in odometry can be tolerated remains an unclear problem. In addition, to meet the accuracy requirement, link positions should also be taken into consideration. \citet{liu2024rdt, he2024learning, he2024omnih2o} add target link positions (in the base frame) as a part of the motion command. With masked selection, moving effectors or feet to the desired position lets the robot perform fine maneuvers. 


\paragraph{Sim-to-Real for Legged Robots}
Due to the over-complication of search and conditioning of traditional model-based planning, reinforcement learning and training in simulation have been widely used recently in control for legged robots. By simplifying the collision shapes and using efficient GPU-accelerated rigid-body physics simulations, quadruped robots~\cite{rudin2022learning, kumar2021rma, escontrela2022adversarial} and humanoid robots~\cite{ilija2024humanoid, liao2024berkeleyhumanoid, xia2024dukehumanoid, cheng2024express, he2024learning, zhang2024wholebody, castillo2021robust} can perform various extreme difficult tasks, such as walking through rough terrain~\cite{nahrendra2023dreamwaq, agarwal2022legged, lee2020learning}, overcoming extreme challenging obstacles~\cite{david2024anymal, zhuang2023robot, cheng2023parkour, zhuang2024humanoid, long2024learninghumanoidlocomotionperceptive}. Training in simulators and zero-shot deployment in the real world seems to be a promising direction for the control algorithm for legged robots. However, these works lay a hidden assumption: only feet are making contact with the hard environment. For humanoid manipulation tasks, the training in simulation does not take hand contacts into consideration~\cite{zhang2024wococo, fu2024humanplus}. When deployed in the real world, the contacts generated by manipulation lead to rigidly attached objects to the robot system, however, it does not involve stochastic contacts of the other parts of the body. These attachments can be directly handled by domain randomization trained in simulation. In this work, we aim to face the contacts directly by training and deploying those contact-rich motions and show that it is possible to train using simplified collision shapes while successfully deploying the policy on the real robot.

\section{Methods and Implementations}
\begin{figure*}
    \centering
    \includegraphics[width=0.8\linewidth]{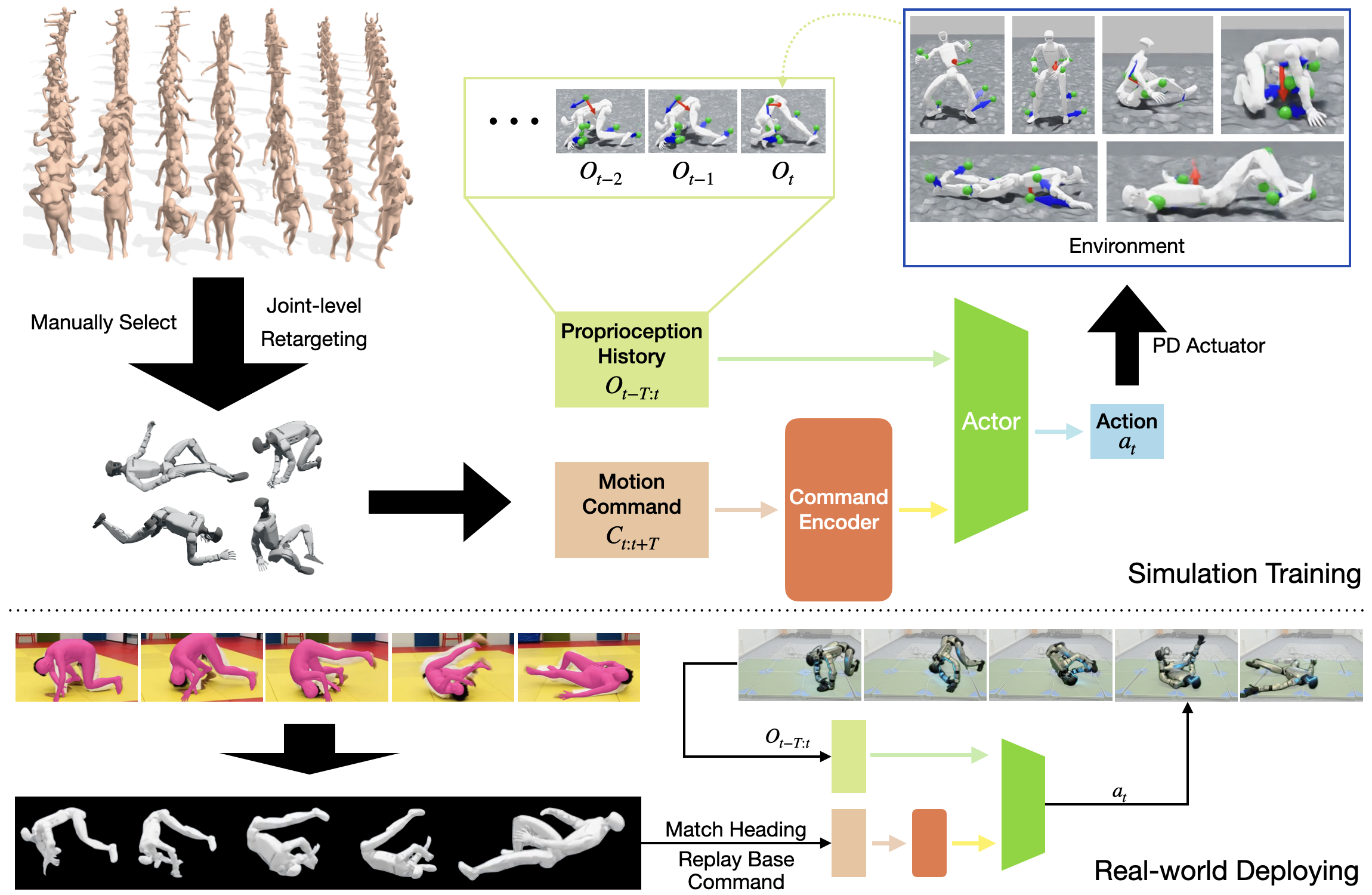}
    \label{fig:training-pipeline}
    \caption{Training Framework: We build an extreme-action dataset from AMASS dataset and internet videos using 4D-Human~\cite{goel2023humans}. We retarget the human motion to the joint-level target of the Unitree G1 robot. We then feed the motion command as a sequence to a Transformer-based encoder. Concatenated with a stack of history proprioception observation (with no linear velocity), we use a sequence of MLP layers to output the joint-level action. In the simulator, a PD controller is used to compute the torque for each joint motor.}
\end{figure*}

\subsection{Addressing the Data Challenge}
\paragraph{Motion commands with multi-contacts}
\begin{figure}
    \centering
    \begin{subfigure}[b]{0.24\textwidth}   
        \centering 
        \includegraphics[width=\textwidth]{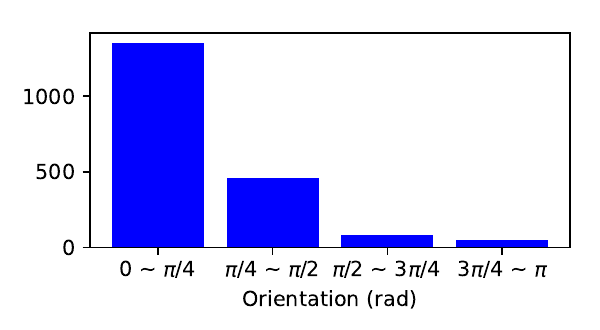}
        \caption[]%
        {{\small The number of files counting maximum base orientation}}    
        \label{fig:orientation-count-by-files}
    \end{subfigure}
    \hfill
    \begin{subfigure}[b]{0.24\textwidth}   
        \centering 
        \includegraphics[width=\textwidth]{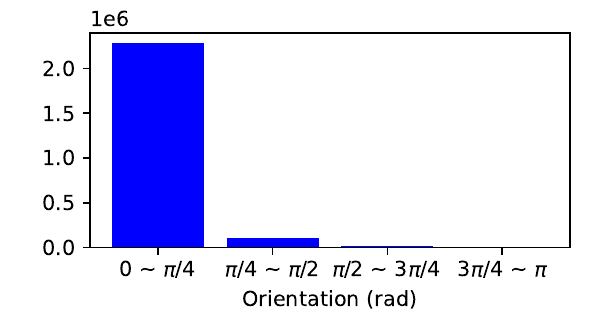}
        \caption[]%
        {{\small The number of frames of base orientation}}    
        \label{fig:orientation-count-by-frames}
    \end{subfigure}
    \vskip\baselineskip
    \begin{subfigure}[b]{0.24\textwidth}
        \centering
        \includegraphics[width=\textwidth]{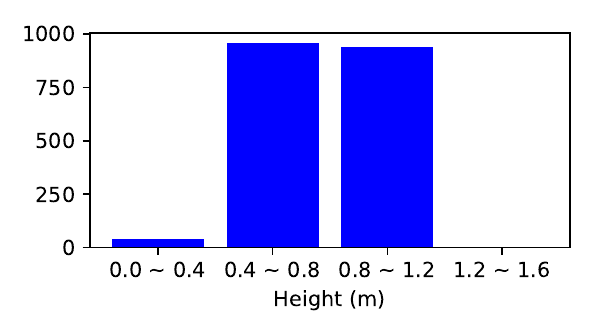}
        \caption[]%
        {{\small The number of files counting smallest base height}}    
        \label{fig:height-count-by-files}
    \end{subfigure}
    \hfill
    \begin{subfigure}[b]{0.24\textwidth}  
        \centering 
        \includegraphics[width=\textwidth]{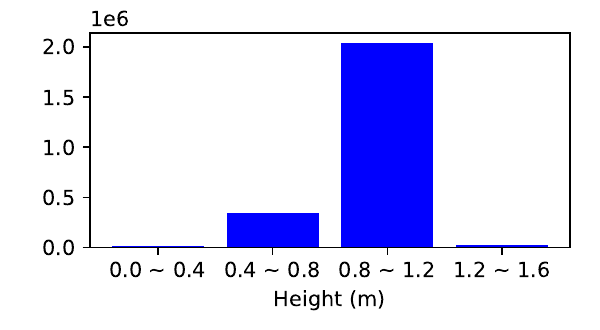}
        \caption[]%
        {{\small The number of frames of base height range}}    
        \label{fig:height-count-by-frames}
    \end{subfigure}
    \caption{Histogram of the number of motions in terms of their maximum roll/pitch and the minimum base height.}
    \label{fig:hist-dataweights}
\end{figure}

To bridge the gap between the high-level commands and the real-time low-level whole-body control algorithm for humanoid robots, a general motion interface must be defined. For both local motions such as motor joint positions and global motions such as the future trajectories of the robot base, a reasonable general motion command is almost impossible to generate from a predefined manifold. Thus, a better option is to sample them from human motion datasets, such as AMASS~\cite{mahmood2019amass}.

However, most of the trajectories in AMASS are collected when the subjects are standing instead of sitting or lying on the ground. As shown in Figure~\ref{fig:hist-dataweights}, we count the number of frames and the number of files whose maximum base orientation and minimum base heights. Figure~\ref{fig:height-count-by-files} counts the number of motion files whose minimum base height reaches a certain range. As shown in Figure~\ref{fig:orientation-count-by-files} and Figure~\ref{fig:orientation-count-by-frames}, most of the motion files are performed when the base position is standing straight. As shown in Figure~\ref{fig:height-count-by-files} and Figure~\ref{fig:height-count-by-frames}, most of the motions are performed when the robot base is over $0.4$m. In this case, not many contact-rich motion is presented in the dataset.
In this case, we build an extreme-action dataset from the AMASS dataset and extract extreme human motions from Internet videos using 4D-Human~\cite{goel2023humans}.

\paragraph{Physical Infeasibility of the Motion Commands}
\begin{figure}
    \centering
    \includegraphics[width=\linewidth]{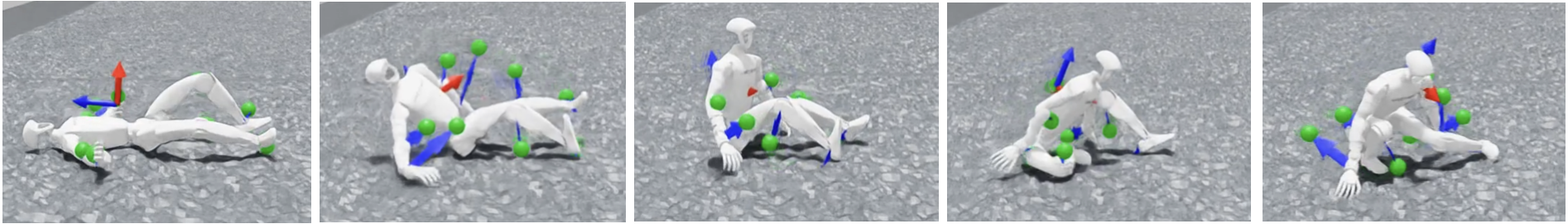}
    \caption{Inconsistency example of humanoid getting up from laying on the ground. The transform frame in the figures are the target base orientation as the motion command.}
    \label{fig:inconsistency-height-plot}
\end{figure}
Even though sampling from humanoid motion commands from a pre-collected human motion capture dataset is a reasonable way to start the training pipeline, the motion trajectory from the dataset still presents inconsistencies after rescaling the size between the human subject and the real humanoid robot. In Figure~\ref{fig:inconsistency-height-plot}, the height trajectory of the motion reference is floating, so the height becomes consistent when the motion reference goes to a standing position. To address this issue, we loose the termination condition which terminates only when the robot is far from the reference trajectory over $0.5$m or the base rotation to the target rotation is greater than $1.0$ rads.

\subsection{Problem Setup}
\label{sec:problem-setup}
\begin{figure}
    \centering
    \includegraphics[width=\linewidth]{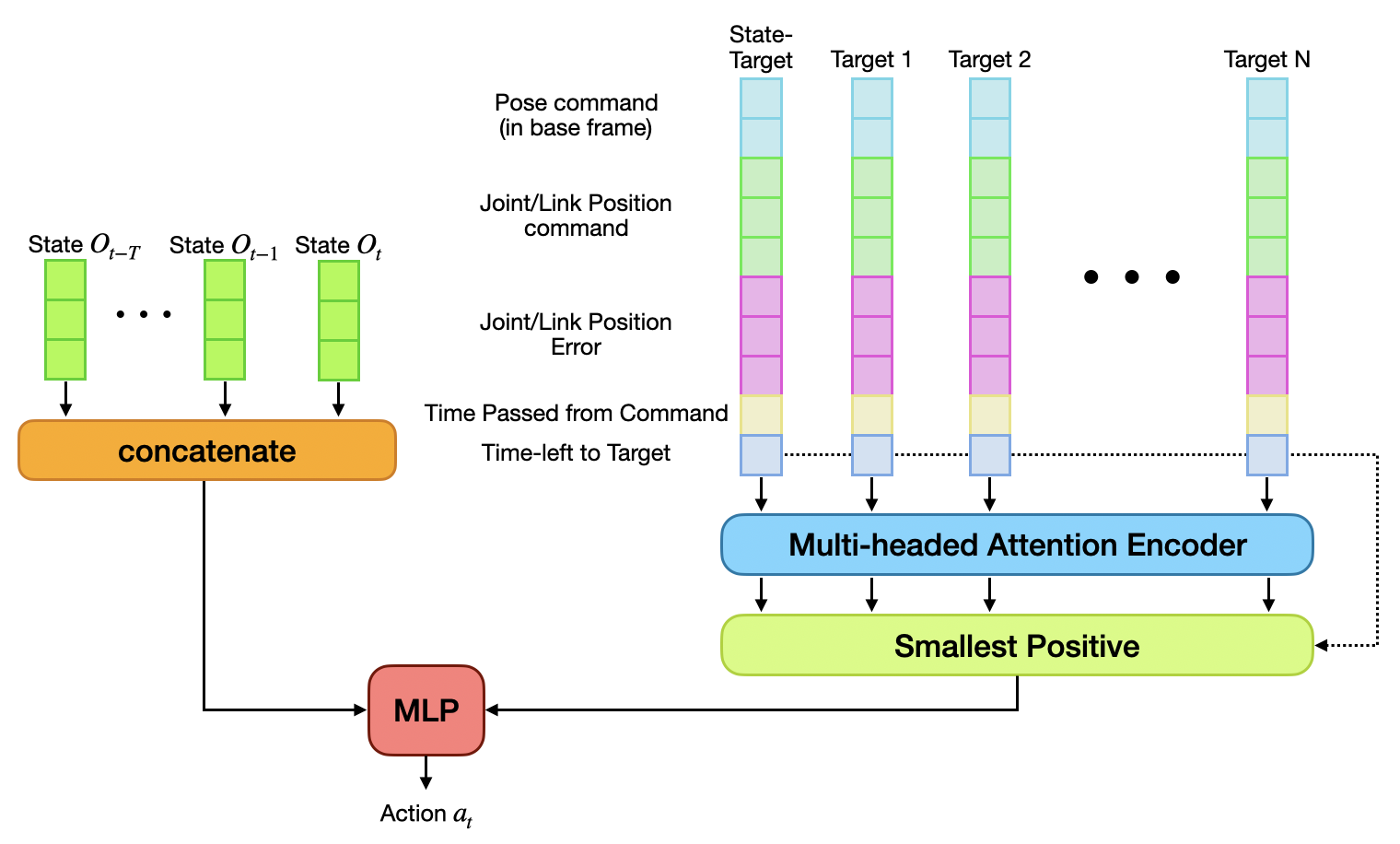}
    \caption{To handle a variable of input motion command, the command encoder adopts a Transformer-based encoder. We select the embedding whose source has the closest ``time-left to target'' value. Then, we concatenate the embedding with the stack of history proprioception and feed it to an MLP to acquire the action output.}
    \label{fig:network-architecture}
    \vspace{-0.4cm}
\end{figure}
We define the task as a low-level humanoid control policy that reaches the target motion in a specified future time. As shown in Figure~\ref{fig:network-architecture}, the motion target is represented in a sequence of concatenated joint positions, target link positions and target base transform under the base when the motion target is refreshed, combined with the time passed from the motion target refreshing and the time to the specific frame. We train the low-level whole-body control policy to reach the target motion in specific frames as close as possible, while not reaching the termination condition.

Since we build an extreme-action dataset by ourselves, we retarget the SMPL format motion data to base position $\hat{p}_w$ and base axis-angle $\hat{\alpha}_w$ in the world frame and the motor joint positions $\hat{\theta}^i$ of the humanoid robot. For each motion command frame, we get the robot base position $p_w$ and the base axis-angle $\alpha_w$. We compute the motion target of the robot base ($\tilde{p}_b$ and $\tilde{\alpha}_b$) under the current robot base frame. We then use forward kinematics to compute the target link positions $\hat{l}_{b}^{j}$ under the robot's base frame. In summary, for each motion target frame, we concatenate all motion targets $[\tilde{p}_b, \tilde{\alpha_b}, \hat{\theta}^i, \hat{l}_{b}^j, (\hat{\theta}^i - \theta^i), (\hat{l}_{b}^j - l_b^j), t_\text{passed}, t_\text{left}]$. $t_\text{passed}$ and $t_\text{left}$ denote the time from the refresh of the motion reference and the time left to this motion target frame respectively. Considering link positions error and joint positions error are both in the local (robot's base) frame, they can be refreshed in real-time. We make these quantities also in real-time in both simulation and real-world deployment.

\subsection{Network and Training Pipeline}
To adapt to the needs of enabling an arbitrary number of motion targets being fed into the network, we use a Transformer-based motion reference encoder. Shown in Figure~\ref{fig:network-architecture}, we add a \textit{state-target} into the motion target sequence to prevent data error when all motion targets are running out. When acquiring the encoded latents, we select the frame that has the smallest positive \textit{time-left-to-target} value. We then concatenate the latent embedding with a stack of history proprioception and feed it into MLP layers to get the joint action as the output.

To train the humanoid motion based on the motion command sequence, while leaving enough flexibility to control policy, we compute the motion target reward only when the frame of motion target is expected to reach. In other words, \textit{time-left-to-target} equals to zero. Therefore, some regularization reward terms must be used. For example the action rate, joint acceleration, energy, joint position out-of-limits, etc. However, these regularization terms are dense while the motion target reward is sparse. The robot's action spikes when the expected motion target is reached if all reward terms are combined. In this case, we use multiple critic networks and perform the advantage mixing technique~\cite{juan2020multicritic} in addition to the PPO algorithm~\cite{schulman2017proximal}.

\subsection{Advantage Mixing for Sparse Task Rewards}
Different from conventional actor-critic architecture, we use one actor network as the policy and 3 critic networks for 3 different groups of rewards $\left(r^{(1)}, r^{(2)}, r^{(3)}\right)$. Each reward groups have multiple reward terms. Following \citet{juan2020multicritic}, each critic network $V_{\Psi^{(i)}}(s_t)$ is supervised independently by their reward group $r^{(i)}$ with temporal difference error,
\begin{equation}
    \mathcal{L}\left(\Psi^{(i)}\right) = \hat{\mathbb{E}}_t \left[\|
        r^{(1)}_t + \gamma V_{\Psi^{(i)}}(s_{t+1}) - V_{\Psi^{(i)}}(s_t)
    \|^2 \right]
\end{equation}
where $\hat{\mathbb{E}}$ is the empirical average and $\gamma$ is the discount factor.
For the policy gradient part, the advantages are combined by weighted average after the general advantage estimation~\cite{schulman2015gae},
\begin{equation}
    \tilde{A} = \sum_{i = 0}^n{w_i \frac{A_i - \mu_{A_i}}{\sigma_{A_i}}}
\end{equation}
where $\{A_i\}_{i=0}^n$ are advantages estimated by the 3 individual critic networks. $\mu_{A_i}$ and $\sigma_{A_i}$ are the batch-wise statistics from each individual reward groups.

\subsection{Termination Condition}
Given most of the humanoid research papers are limited to the standing orientation for the base frame, the termination condition is easy to define. \citet{zhuang2024humanoid, long2024learninghumanoidlocomotionperceptive, cheng2024express, ilija2024humanoid, he2024learning, he2024omnih2o} all fit the termination condition when the robot's base has large roll/pitch angle or has a height below a certain threshold. However, interacting with the ground, such as crawling and spinning, intrinsically makes the robot's height significantly lower than the standing pose. Lying-down on the ground is a ``fall'' state under the context of standing. In this case, we define the termination condition by the difference between the humanoid robot and the target motion. We apply the termination condition only when the motion command is expected to be achieved at a certain timestep. A training rollout is terminated if any of the following conditions are met at the expected reach time,
\begin{equation}
    \|p_w - \hat{p}_w\|_2 > 0.5
\end{equation}
where $p_w$ and $\hat{p}_w$ are the base position of the robot and the base position of the motion command in the world frame,
\begin{equation}
    \| \text{Im} (\text{inv}(q_w) * \hat{q}_w) \| < 0.8
\end{equation}
where $\text{Im}(\cdot)$ is the imaginary part of the quaternion, $q_w$ and $\hat{q}_w$ is the quaternion of the robot base and the motion command in the world frame,
\begin{equation}
    \forall j \in \text{N}, |\theta^j - \hat{\theta}^j| > 1.0
\end{equation}
where $\text{N}$ is the number of joints on the robot, $\theta^j$ and $\hat{\theta}^j$ are the joint positions on the robot and the motion commands respectively.

\subsection{Deployment}
\begin{figure}[ht!]
    \centering
    \includegraphics[width=0.8\linewidth]{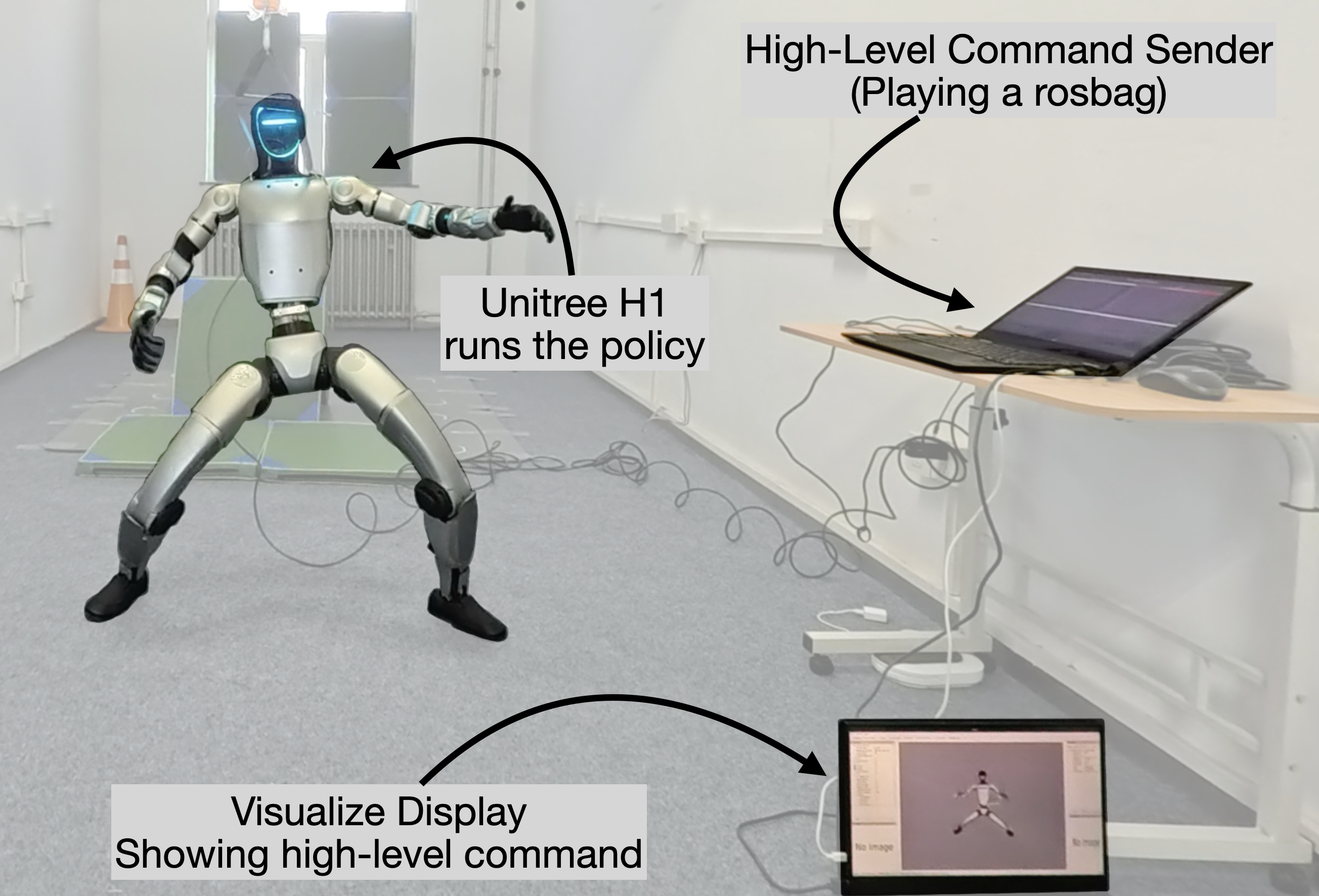}
    \caption{We run the control policy (including the Transformer-based encoder) using Nvidia Jetson NX inside of Unitree G1. We use an additional laptop to serve as another hardware that sends high-level motion commands, which are visualized in the bottom right of the figure. No motion capture system is used.}
    \label{fig:hardware-architecture}
    \vspace{-0.6cm}
\end{figure}
One of the key statements of this work is to train a deployable sim-to-real controller to perform contact-agnostic motion. We deploy the policy onboard to test the entire pipeline. Our system is designed as a low-level controller commanded by a high-level motion generator. We set up another laptop as shown in Figure~\ref{fig:hardware-architecture}, we connect the laptop to the Unitree G1 Robot by Ethernet and ROS2. We then use this laptop to replay a pre-recorded target motion command sequence, which also testifies the communication latency through Ethernet and ROS2. But to notice that, the low-level controller policy is still running on the Nvidia Jetson Orin inside of the Unitree G1's torso. We use ROS2 messages to communicate with the hardware onboard, such as motors and IMU. We use ONNX~\cite{onnxruntime} to accelerate the computation of the policy network.

Position errors $(\hat{l}_b^j - l_b^j)$ for the target links are computed via Pytorch Kinematics~\cite{Zhong2024PyTorchKinematics} using the forward kinematics. When deploying onboard, the forward kinematics is exported as an ONNX program since the entire exporting process is tracking the PyTorch computation graph and turning all constants, such as fixed transform between links, into the exported program file. The forward kinematics for the Unitree G1 and for the specific target links can run onboard with ONNX acceleration.

We hypothesize that odometry is not a dominant factor towards the success of these contact-agnostic motions. We play the base position targets under the base frame when recorded in the simulator rollouts. Specifically, we play the will-trained low-level control policy in the simulator and record the target joint positions and target link positions into a rosbag file. In the meantime, we record the target base position with respect to the base frame in the simulator. We also record the target base orientation in the world frame and compute the target base orientation with respect to the robot's base frame onboard, because we have access to the IMU data in the world frame in realtime during deployment.

But computing the target base orientation with respect to the robot's base frame in real time leads to another issue: The heading of the motion command is typically not aligned with the robot base heading. Due to the issue of gimbal lock, extracting the yaw components of the orientation of both motion command and robot's IMU does not solve this issue. At the start of the motion, we put the robot in the joint positions of the first frame of the motion command. We then put the robot on the ground, for example laying or facing down, getting the quaternion $q_\text{ref}$ of the target motion command and the quaternion $q_\text{robot}$ of the IMU reading. Then, we extract the component of $q_\text{correct} = q_\text{robot} * \text{conj}(q_\text{ref})$ by
\begin{equation}
    q_\text{correct} = (w_c, x_c, y_c, z_c)
\end{equation}
\begin{equation}
    \gamma_c = \text{atan}(2 * (w_c * z_c + x_c * y_c), 1 - 2 * (y_c^2 + z_c^2))
\end{equation}
\begin{equation}
    q_\text{yaw,correct} = (\cos{(\gamma_c / 2)}, 0, 0, \sin{(\gamma_c / 2)})
\end{equation}
During the replay of the target motion command, each target base orientation in the global frame is left-multiplied with this $q_\text{yaw,correct}$.

\section{Experiments}
\begin{table*}[ht!]
\centering
\begin{tabular}{c|ccc}
\multicolumn{1}{l|}{}                                                                 & \textbf{Get up from ground} & \textbf{Ground Interaction} & \textbf{Standing Dance} \\ \hline
\textbf{\begin{tabular}[c]{@{}c@{}}Multi-Critic\\ \& Selected Motions\end{tabular}}   & 94.30\%                     & 98.25\%                                          & 100\%                   \\ \hline
\textbf{\begin{tabular}[c]{@{}c@{}}Single-Critic\\ \& Selection Motions\end{tabular}} & 65.13\%                     & 84.12\%                                          & 100\%                   \\ \hline
\textbf{\begin{tabular}[c]{@{}c@{}}Multi-Critic\\ \& AMASS Motions\end{tabular}}      & 1.45\%                      & 46.56\%                                          & 100\%                  
\end{tabular}
\caption{Success rate under the same number of iterations and the same reward configurations.}
\label{tab:success-rate}
\vspace{-0.5cm}
\end{table*}
In this section, we set up 3 experiments to verify the effectiveness of the designs and components of our method. We run quantitative results in the simulator. First, we show the effectiveness of the multi-critic, a.k.a advantage mixing technique, through a specific case study of the robot behavior. Then, we explain through the robot's behavior why we have to train using an extreme-action dataset to address the significance of training a sim-to-real-possible motion in the simulator.

To verify the successful deployment of our proposed pipeline, we build the extreme-action dataset with 3 types of motions. 2 of them are some extremely difficult motions because they introduce unexpected contacts between robot parts and the ground.
\paragraph{Get up from ground} The first type of motion is getting up from ground, such as subject 140 CMU mocap~\cite{cmumocap} and subject 3 in KIT~\cite{Mandery2016unifying}.
\paragraph{Ground interactions} Another type of motion is ground interaction motion. We collect the motion example from internet videos and track the human motion using 4D-Human~\cite{goel2023humans}. We then retarget these human motion in SMPL format to the joint positions of the Unitree G1 robot. 
\paragraph{Standing Dance} The last type of motion is mostly standing and dancing. We also testify our pipeline by tracking a sequence of human motion from internet video using 4D-Human~\cite{goel2023humans} and retarget the motion to the humanoid robot and play in the simulator.

\begin{figure*}[ht!]
    \centering
    \includegraphics[width=0.8\linewidth]{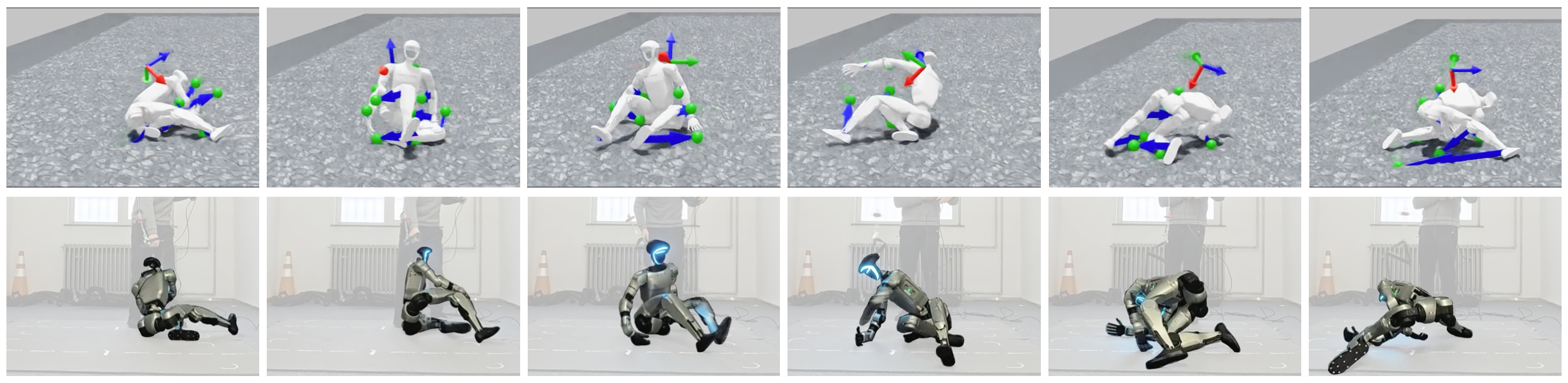}
    \caption{We select the most challenging task where the robot is doing a breaking dance move on the ground. The hip, knee, thigh, elbow, and hands are contacting the ground expected or unexpectedly. To be noted that, the rubber hands on the robot cannot be simulated in the simulator, we use the rigid-body collision shape by making a convex hull of the hands' mesh.}
    \label{fig:groundspin-analysis}
    \vspace{-0.5cm}
\end{figure*}

We test the success rate of these 3 types of motions by generating 1000 robots with uniformly sampled domain randomization parameters and run 10 times in the simulation. We count the number of trajectories being run and the number of trajectories being terminated because of the failure.

\subsection{Effectiveness of Multi-Critic}
Due to the discrete feature of motion command reward, mixing regularization reward terms together with the motion (task) reward disrupts policy learning. We compare the learning curve of using multi-critic and single-critic techniques. Due to the in-feasibility of those collected motion commands, finishing these motions is the top priority, not the mean per joint error. In terms of single-critic, we add the task reward together with the regularization reward as well as the safety reward and make a single reward function.
Using the multi-critic technique leads to faster training speed and faster convergence rate. Single-critic method indeed leads to similar performance, but significantly lowers the training and convergence speed, which reduces the experiment efficiency. Shown in Table~\ref{tab:success-rate}, training using the multi-critic method leads to a higher success rate in either of the contact-agnostic tasks.

\begin{figure}[ht!]
    \centering
    \includegraphics[width=0.9\linewidth]{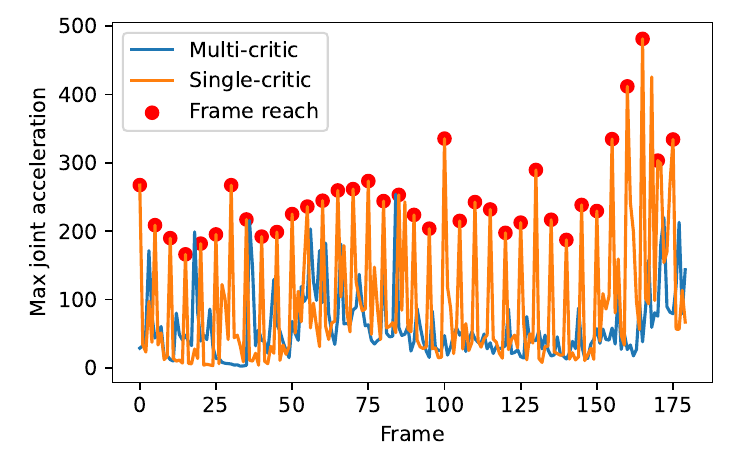}
    \caption{Maximum joint acceleration reported when playing the ``get-up-from-laying'' (CMU-140, 140\_08) motion in simulator.}
    \label{fig:joint-acc}
    \vspace{-0.6cm}
\end{figure}

We also do a detailed behavior analysis through one case study. We play the behavior of training using multi-critic and single-critic. For example, getting up from lying on the ground, as shown in Figure~\ref{fig:inconsistency-height-plot}, the robot has to follow the motion command and get up from the laying pose. Shown in Figure~\ref{fig:joint-acc}, the red dots indicate the timestep when the motion command is expected to be reached. The maximum joint acceleration plot of the single-critic-trained policy always generates a spike as the policy is trying its best to reach the commanded motion. However the policy trained on the multi-critic method has a lot smoother behavior, with no obvious joint acceleration spike. We encourage readers to take a look at the supplementary video for more details.

The reason why we don't report the average joint error or average link position error is that these extreme motions for the current robot is not physically feasible. The policy needs to learn trade-offs between not being terminated to reach more targets and following the joint position command more accurately.

\subsection{Importance of using an Extreme-Action dataset}
\begin{figure}[ht!]
    \centering
    \includegraphics[width=0.8\linewidth]{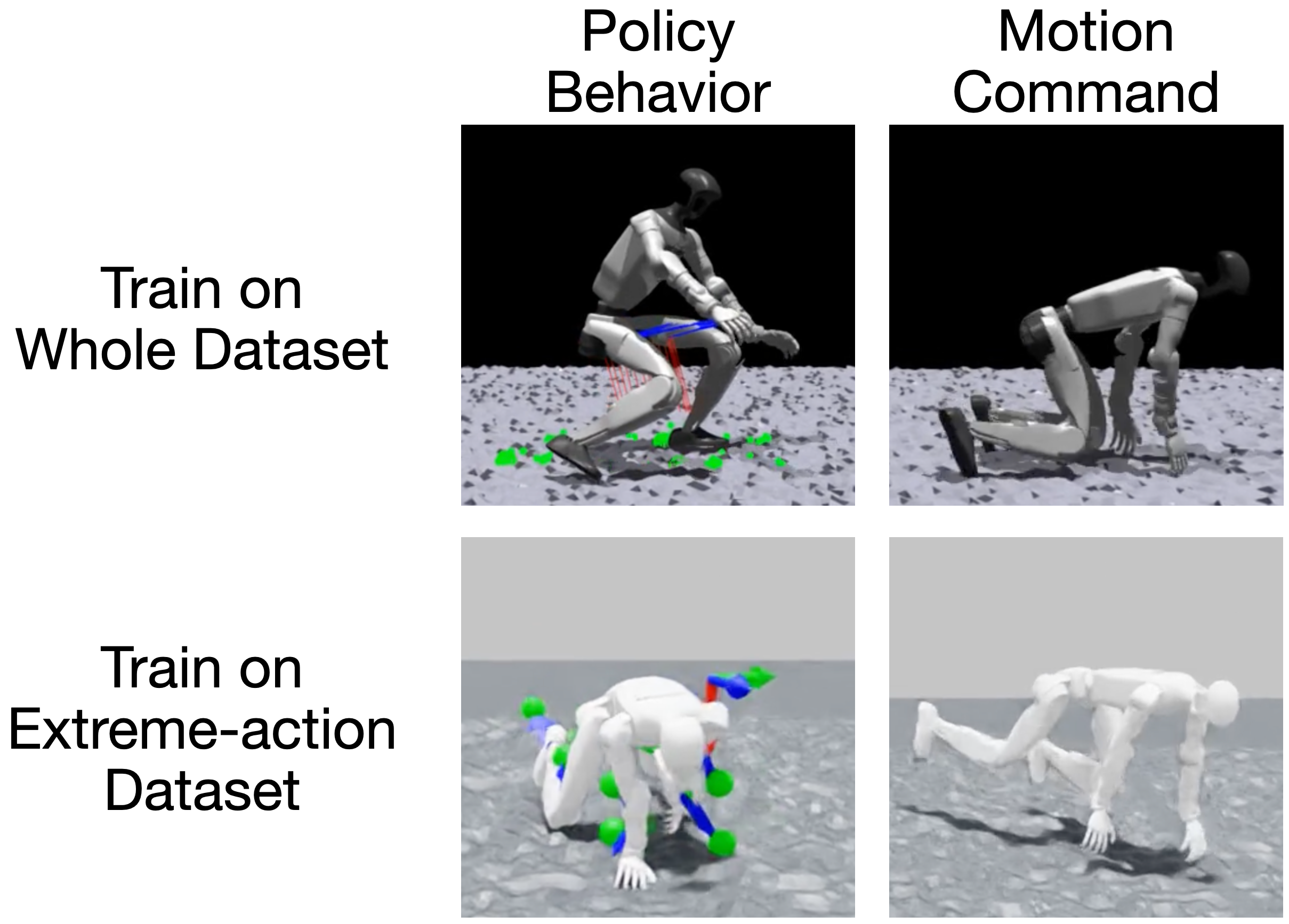}
    \caption{The comparison between to policy behaviors of crawling from a standing pose, using the AMASS dataset and the extreme-action dataset.}
    \label{fig:why-extreme-motion}
    \vspace{-0.6cm}
\end{figure}
In this section, we explain through a specific case study why we only select a handful of motions to train and achieve these extremely difficult motions in the real humanoid robot. As shown in Figure~\ref{fig:why-extreme-motion}, if we train on the entire AMASS dataset, the lowest position the robot can reach is squatting. As shown in Figure~\ref{fig:hist-dataweights}, even with 4096 parallel robot rollouts during training, the motion commands of contacting with the ground are not trained enough under a reasonable experiment iteration. Thus, data bias leads to the ``safer'' motion for the policy, which refuses to contact the ground using its hands. Since the main focus of this work is to train a deployable policy that performs extreme humanoid motion in the real world, generalization ability is not the primary concern. We trained on an extreme-action dataset, which is described in supplementary in detail. Also, shown in Table~\ref{tab:success-rate}, the policy trained on the AMASS dataset still only learns the behavior when standing, but does not successfully react to contact-agnostic motion command. However, our method still presents generalization ability when training on standing motions. The detailed humanoid behavior will be presented in the supplementary video.

\subsection{Behavior Analysis of Extreme Difficult Contact-Agnostic Motions}
We select an extreme motion to show the difficulty and the successful deployment of our method. Shown in Figure~\ref{fig:groundspin-analysis}, we choose a short motion of breaking dance video from the internet. The motion is to sit on the ground and get up by turning to another direction. This motion involves the collaboration using hands, hips and legs. At the stand up phase, the robot needs to use its knee to raise its body. Especially in the last image of Figure~\ref{fig:groundspin-analysis}, the robot in the simulator is able to get up with only its leg and knee. The robot in the real world has a balance problem and has to use the hands. In addition, in the middle stage of this motion, the robot needs to use its hand to keep balance. But hands are rigid in the simulator and deformable in the real world.

Also, we observe that future motion target indeed help our policy plan ahead, as well as increase the success rate of the whole motion sequence. In the example of ``get-up-from-ground'', the robot starts by lying on the ground. The robot fails to shift its center-of-mass forward in preparation of getting up, if the policy is fed with only one immediate motion command. The success rate significantly increases as future command is included. More details are described in the supplementary.

\section{Conclusion} 
\label{sec:conclusion}
In this work, we embrace collisions that have almost never been discussed in recent humanoid research. We overcome the exponential searching issue in model-based control for humanoid robots using zero-shot sim-to-real reinforcement learning. We propose a general motion command for humanoid so that locomotion, manipulation and whole-body control tasks can be unified in a single interface. Based on this motion command, we adopt a transformer-based encoder to process the command input with a variable input length. By diving deep into the motions where humanoids contact with the environments with components not limited to hands and feet, we show the potential of training in GPU accelerated simulation using Reinforcement Learning can make such complex and extremely difficult motion realizable in the real world, even if the motion command is not physically feasible for the given robot model.

\noindent\textit{Limitations:} Even though this work shows the potential of training complex motions only in a simulator and making them possible in the real world, training a low-level controller that performs full humanoid motion still requires a high-quality motion dataset that is not only limited to standing motions. Or we need a way of bridging the gap between robot models and human subjects among these collected human motion datasets. For high-level commands, for example, large action models, they do not consider leg motions. We need to build an abstraction such as masking on the lower legs command to make the entire general humanoid motion system possible in the future.
Addressing these limitations and training a general humanoid whole-body control system that allows a high-level large action model to reason and send general motion commands to the real humanoid robots will be our future work.

\section*{Acknowledgments}

\bibliographystyle{plainnat}
\bibliography{references}

\end{document}


\title{\textbf{Supplementary for}: Embrace Collisions: Humanoid Shadowing for Deployable Contact-Agnostics Motions}

\author{Author Names Omitted for Anonymous Review. Paper-ID [464]}

\maketitle

\begin{figure*}[ht!]
    \centering
    \includegraphics[width=0.8\linewidth]{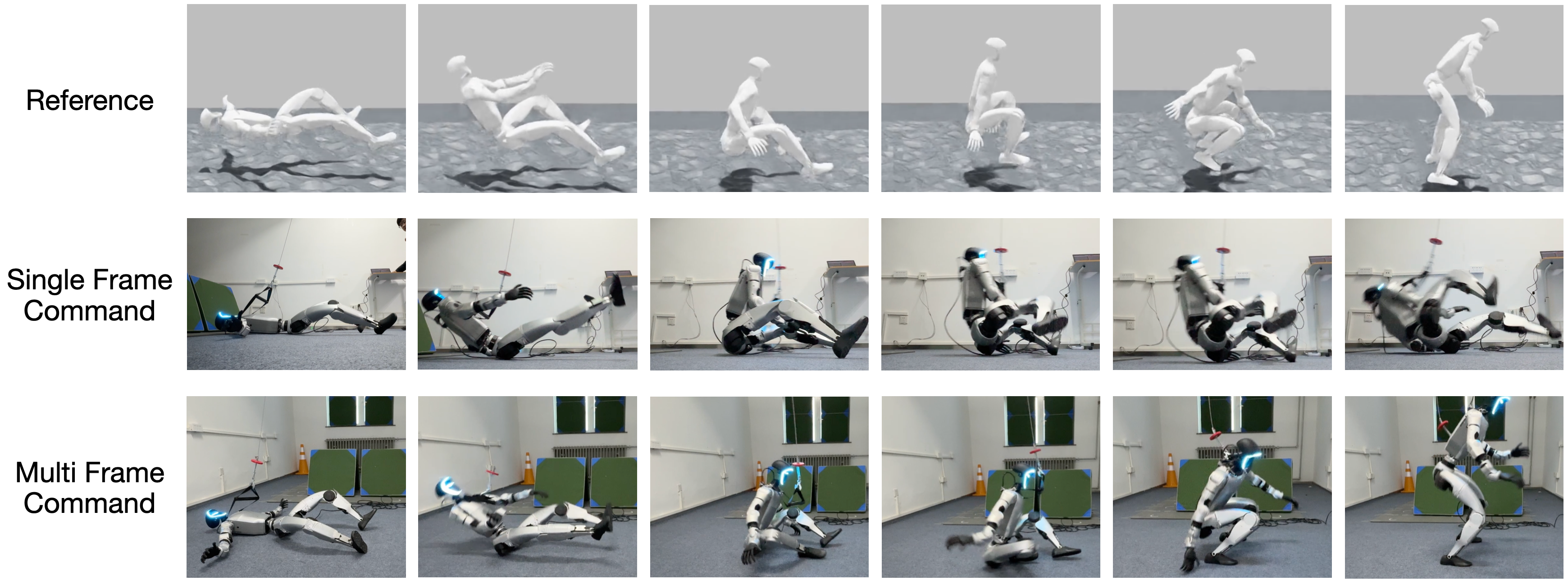}
    \caption{Comparison between multiple frames of motion command and single frame of motion command, in the real world.}
    \label{fig:multi-motion-command}
\end{figure*}
\begin{figure*}
    \centering
    \includegraphics[width=0.8\linewidth]{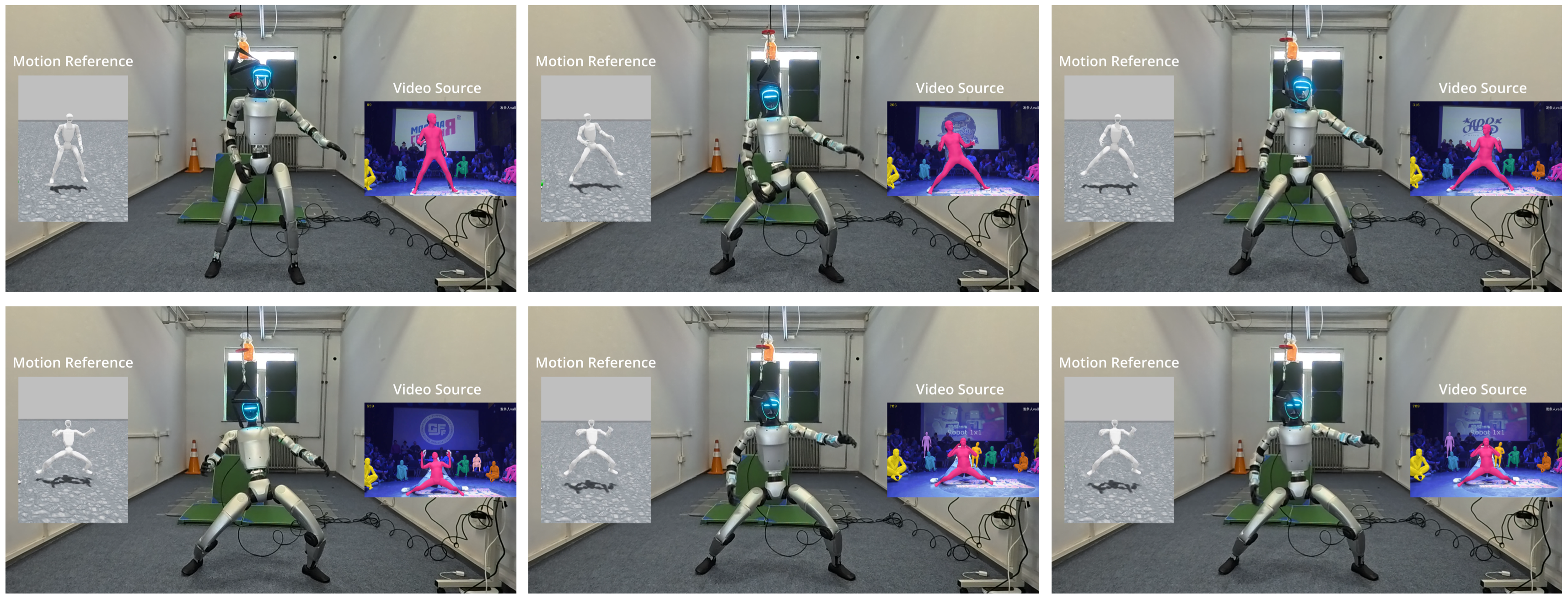}
    \caption{Video snapshots for running Kpop-dance movement in the real-world.}
    \label{fig:kpop-dance}
\end{figure*}

\section{Video}
We attach a video describing the main idea and the real-world experiments in real-time with no accelerations. We perform real-world tests on all three types of motions. We encourage readers to see the video for a more comprehensive understanding of this work.

\section{More Results}

\begin{figure}[ht!]
    \centering
    \includegraphics[width=0.8\linewidth]{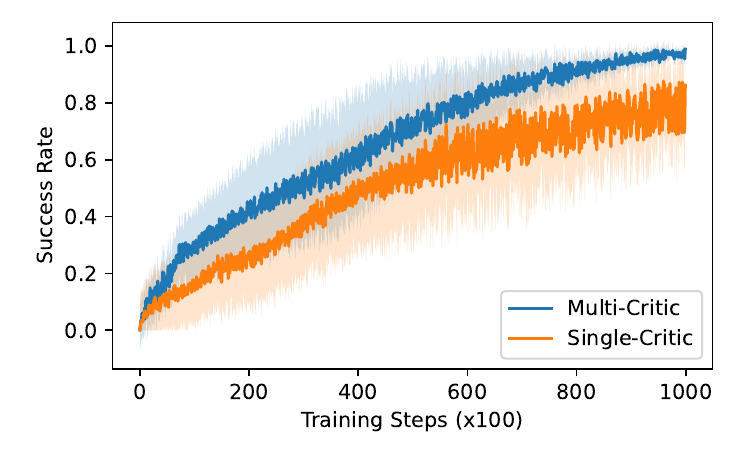}
    \caption{The success rate reported during training when trained on all 4 extreme difficult contact-agnostic motion.}
    \label{fig:training-success-rate}
\end{figure}
Figure~\ref{fig:training-success-rate} shows the training progress comparing multi-critic technique and single-critic technique. Using multi-critic setting leads to faster training speed as well as faster convergence rate.

We also do ablation study on why a sequence of motion command is necessary. We run the exact same policy onboard, but provide different length of motion command to the robot, shown in Figure~\ref{fig:multi-motion-command} The upper row presents the motion reference shown in simulator, which is also the immediate motion command to be achieved. The middle row presents the real-world robot's snapshots where only the immediate motion command is sent to the low-level policy. At current setting, the low-level policy only gets one frame of motion command. In the bottom row, the low-level policy receives a sequence of motion command. As the Figure~\ref{fig:multi-motion-command} shows, if future command is feed to the low-level policy, it presents preparing behavior to shift it's weight a bit forward to for the future standing behavior.

\section{Simulation Details}
As described in the main text, we select a handful of motion commands to train the humanoid whole-body controller to overcome the unbalanced issue of the precollected dataset.
\begin{table}[ht!]
    \centering
    \begin{tabular}{c|cc}
        \textbf{Dataset}        & \textbf{Subject}                                                                  & \textbf{Motion names}            \\ \hline
        CMU            & 140                                                                      & Get up from ground      \\ \hline
        KIT            & 3                                                                        & Crawling                \\ \hline
        Internet Video & {Bilibili BV1L34y1t71x} & Popping dance movement  \\ \hline
        Internet Video          & {BiliBili BV1Nm4y1k7wP} & Breaking dance movement \\ \hline
        Internet Video          & {Bilibili BV1mv411G7WM} & Jiu-jitsu movement     
    \end{tabular}
    \caption{The motion reference we select to build tine motion command dataset.}
    \label{tab:data-selection}
\end{table}

Shown in Table~\ref{tab:data-selection}, we train our policy in simulation by extracting motion command from these motion references.

Considering these extreme difficult humanoid motion reference data may lead to physically infeasible situations, such as body parts penetrating the ground or robot base floating in the air, we set the robot's joint positions as the retarget robot pose in the first motion reference frame. We initialize the robot's base position by first adding a positive height to all motion reference frame so that no motion reference frame is penetrating the ground. Then we add a tiny height offset to spawn the robot, typically 0.06m. Also, to help the policy experience more states if the policy stuck as some place, for example if it does not get up from the ground, we sample the initial pose of the robot not only from the first frame of the motion reference trajectory, but from the start of the motion reference to the $60\%$ of it.

During each rollout, we select one motion reference to generate the motion command. At time $t$, we use a pre-sampled time interval $t_\text{int}$ and sample the motion reference at $t + t_\text{int}, t + 2t_\text{int}, t + 3t_\text{int}, t + 4t_\text{int}, t + 5t_\text{int}$ respectively. We also compute the base reference position and orientation in the base frame at time $t$.

\begin{table*}[ht!]
    \centering
    \begin{tabular}{l|ll}
                                              & \textbf{Names}                           & \textbf{Value}     \\
                                              \hline
        Environments                          & Number of robots                         & 4096               \\
                                                \hline
        \multirow{5}{*}{Domain randomization} & Scaling body mass                        & 0.8 $\sim$1.2      \\
                                              & Center of mass position                  & -0.02m $\sim$0.02m \\
                                              & Scaling motor stiffness                  & 0.9 $\sim$1.1      \\
                                              & Scaling motor damping                    & 0.9 $\sim$1.1      \\
                                              & Motor delays                             & 0.0s $\sim$0.03s   \\
                                              \hline
        \multirow{2}{*}{Initialize pose}      & Height offset                            & 0.04m              \\
                                              & Sampling frame ratio from the trajectory & 0.0 $\sim$0.6     
    \end{tabular}
    \caption{Detailed parameters for running the system in the simulator}
    \label{tab:domain-rand}
\end{table*}

\section{Training Details}

\begin{table*}[ht!]
    \centering
    \begin{tabular}{c|cl}
        Reward group                    & Reward term               & \multicolumn{1}{c}{Expression} \\ \hline
        \multirow{3}{*}{Task}           & Base position tracking    & $\Psi(\Delta(p_b), 0.4)$  \\
                                        & Base orientation tracking & $\Psi(\Delta(q_b), 0.8)$  \\
                                        & Joint position tracking   & $\Psi(\|\theta^j - \hat{\theta}^j\|, 0.3)$ \\ \hline
        \multirow{3}{*}{Regularization} & Action rate               & $\Psi(\|a^j_t - a^j_{t-1}\|, 1.0)$ \\
                                        & Joint acceleration        & $\Psi(\|\ddot{\theta}^j\|, 500)$ \\
                                        & Joint velocity            & $\Psi(\|\dot{\theta}^j\|, 15)$ \\ \hline
        \multirow{2}{*}{Safety}         & Joint position limit      & $\Psi(\max{(\theta^j - \theta^j_\text{max}, \theta^j_\text{min} - \theta^j)}, 0.1)$ \\
                                        & Joint torque limit        & $\Psi(\max{(|\tau^j| - 0.9~\tau^j_\text{max}, 0)}, 0.1)$
    \end{tabular}
    \caption{Reward terms and their expressions}
    \label{tab:rewards}
\end{table*}
In Table~\ref{tab:rewards}, the function $\Psi$ is a Gaussian kernel where,
\begin{equation}
    \Psi(a, b) = \exp{(-a / b^2)}
\end{equation}
Shown in Table~\ref{tab:rewards}, we build these reward functions in the range of $0 ~ 1$ so that everything is positive, potentially preventing active termination behavior. Then we \textbf{multiply} all reward terms in each reward group so that the algorithm will not completely ignore any of these terms. For the experiment variant using single critic, the reward terms within each group are multiplied and the reward groups are sum together weighted by the same weight parameters of the advantage mixing to get the scalar reward.

\begin{table}[ht!]
    \centering
    \begin{tabular}{c|c}
        Hyperparameters          & Value               \\ \hline
        Optimizer                & AdamW               \\
        $\beta_1, \beta_2$       & 0.9, 0.999           \\
        Learning rate            & $1e-4$              \\
        Batch size               & 4096                \\
        Clip param               & 0.2                 \\
        Entropy coefficient      & 0                   \\
        min\_std clip            & 0.2                 \\
        Desired KL               & 0.01                \\
        Maximum gradient norm    & 1                   \\
        Num minibatches          & 4                   \\
        $\gamma$                 & 0.99                \\
        $\lambda$                & 0.95                \\
        Advantage mixing weights & [0.7, 0.1, 0.2]
    \end{tabular}
    \caption{Parameters in Algorithm implementation}
    \label{tab:ppo-param}
\end{table}

\begin{table}[ht!]
    \centering
    \begin{tabular}{c|c}
        Hyperparameters               & Value               \\ \hline
        Encoder Activation            & GELU                \\
        Encoder Project Activation    & ReLU                \\
        Encoder num heads             & 1                   \\
        Encoder num layers            & 2                   \\
        Encoder d\_model              & 128                 \\
        Encoder feedforward dimension & 128                 \\
        Encoder output size           & 128                 \\
        MLP hidden sizes              & [512, 256, 256] \\
        MLP Activation                & ELU                
    \end{tabular}
    \caption{The detailed network parameters for the low-level policy, which runs onboard}
    \label{tab:network}
\end{table}
The policy network consist of actor and multiple critics with the same structure. We use a transformer-based encoder block to encoder all motion command. The encoder outputs a sequence of embedding, which we select the embedding whose `time-to-target` attribute is the smallest positive value. We then concatenate this embedding with a stacked history proprioception observation and feed them to a Multi-Layer Perceptron. The MLP layers outputs the 29-dof action as the target position to the robot motors. Detailed parameters for the network are shown in Table~\ref{tab:network}.

We train our algorithm on a Nvidia 4090D GPU with 4096 robots in parallel for about 72 hours from scratch. We build the simulation environment using IsaacLab and modify the reinforcement learning framework based on rsl\_rl.

\section{Deployment and Real-World Experiment Details}
\begin{table}[ht!]
    \centering
    \begin{tabular}{c|cc}
        \textbf{Joint name}                & \textbf{Stiffness (kp)} & \textbf{Damping (kd)} \\ \hline
        Left/right shoulder pitch/roll/yaw & 25                      & 1.0                   \\
        Left/right elbow                   & 25                      & 1.0                   \\
        Left/right wrist roll              & 25                      & 1.0                   \\
        Left/right wrist pitch/yaw         & 5                       & 0.5                   \\
        Waist roll/pitch                   & 60                      & 2.5                   \\
        Waist yaw                          & 90                      & 2.5                   \\
        Left/right hip pitch/roll/yaw      & 90                      & 2.0                   \\
        Left/right knee                    & 140                     & 2.5                   \\
        Left/right ankle pitch/roll        & 20                      & 1.0                  
    \end{tabular}
    \caption{Parameters that runs on the hardware}
    \label{tab:hardware}
\end{table}

To run the trained policy on the real robot, we deploy the entire system on an Nvidia Jetson Orin NX and a laptop running Intel i5 CPU. We export the policy (including the transformer-based encoder) as an ONNX program. All components communicate using ROS2 in the network of Unitree G1 robot. We then run the policy on the Jetson board at 50Hz. Since the policy outputs the action as the target joint position of each motor on the robot, we use the built-in PD controller on the Unitree G1 robot, which runs at 1000Hz, with the kp/kd setting as shown in Table~\ref{tab:hardware}. These kp/kd parameters are also the same when training in simulator.

To acquire the target link position and their error respectively, we use Pytorch\_Kinematics~\cite{Zhong2024PyTorchKinematics} and ONNX~\cite{onnxruntime} to export the forward kinematics computation as an ONNX program. The exported ONNX program gets the joint positions and outputs the target link positions in the robot's base frame, which runs in real time on Nvidia Jetson Orin NX.

Since this work is also a proof-of-concept for building a hierarchical general humanoid controller, with a low-level whole-body control policy and a high-level command sender, we use another laptop to send the high-level command which simultaneously test the communication latency. Considering our high-level motion command is defined with base pose sequence under the robot frame when the command is generated, the high-level motion command for the real-world testing cannot be played directly from SMPL-based motion file. We play each motion in simulation using the well-trained low-level policy and record the motion command, as well as the base pose command under the robot's base frame in simulation. We then play this base pose command in the real world and ignore the difference between the robot trajectory in the simulator and in the real world.

In the real-world testing, it is important to determine whether the testing motion succeeded, while we don't install additional motion capture system. For each extreme motion, we determine the success of each motion as finishing the entire motion command sequence with no unexpected head contacting the ground. For getting-up-from-ground task, we terminate the test when the robot's torso orientation significantly deviate from the motion command. In our real-world experiment, we also visualize the motion command in the laptop that sends the motion command sequence.

\bibliographystyle{plainnat}
\bibliography{references}